\documentclass[final]{cvpr}

\usepackage{times}
\usepackage{epsfig}
\usepackage{graphicx}
\usepackage{amsmath}
\usepackage{amssymb}

\usepackage[pagebackref=true,breaklinks=true,colorlinks,bookmarks=false]{hyperref}

\def\cvprPaperID{16} 
\def\confYear{CVPR 2021}

\begin{document}

\title{The Fishnet Open Images Database: A Dataset for Fish Detection and Fine-Grained Categorization in Fisheries}

\author{Justin Kay\\
Ai.Fish\\
{\tt\small justin@ai.fish}
\and
Matt Merrifield\\
The Nature Conservancy\\
{\tt\small mmerrifield@tnc.org}
}

\maketitle

\begin{abstract}
Camera-based electronic monitoring (EM) systems are increasingly being deployed onboard commercial fishing vessels to collect essential data for fisheries management and regulation. These systems generate large quantities of video data which must be reviewed on land by human experts. Computer vision can assist this process by automatically detecting and classifying fish species, however the lack of existing public data in this domain has hindered progress. To address this, we present the Fishnet Open Images Database, a large dataset of EM imagery for fish detection and fine-grained categorization onboard commercial fishing vessels. The dataset consists of 86,029 images containing 34 object classes, making it the largest and most diverse public dataset of fisheries EM imagery to-date. It includes many of the characteristic challenges of EM data: visual similarity between species, skewed class distributions, harsh weather conditions, and chaotic crew activity. We evaluate the performance of existing detection and classification algorithms and demonstrate that the dataset can serve as a challenging benchmark for development of computer vision algorithms in fisheries. The dataset is available at \url{https://www.fishnet.ai/}.

\end{abstract}

\section{Introduction}

More than a thousand commercial fishing boats carry electronic monitoring (EM) systems, which use onboard cameras to track fishing activity and provide accountability in the global seafood market.  These systems produce large volumes of video data that are screened by trained human reviewers. Over the next ten years, the number of boats carrying electronic monitoring systems is expected to grow by 10-20x, outpacing current review capacity \cite{catalyzing}. Computer vision has the potential to drastically reduce the time required to analyze this video by automatically detecting and classifying fish. However, the lack of publicly available data in this domain, and the systemic barriers to obtaining it, have hindered progress on these tasks.

\begin{figure}[t]
\begin{center}
\includegraphics[width=1.0\linewidth]{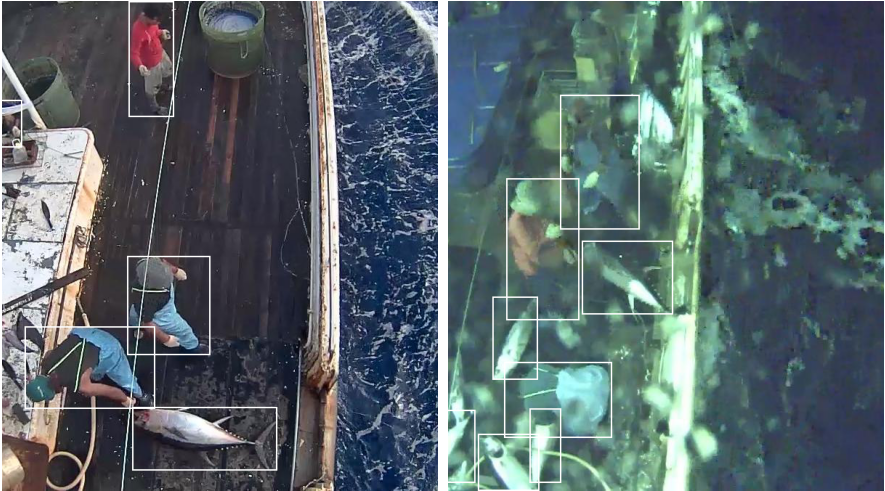}
   \caption{Two example images from the Fishnet dataset, captured by EM cameras mounted above the deck on longline tuna vessels. (Left) Ideal conditions: good lighting, high visibility, and no occlusion. (Right) Challenging conditions: low visibility due to harsh weather, water on the lens, artificial lighting, and hectic crew activity which occludes the fish.}
\end{center}
\label{fig:1}
\end{figure}

\begin{figure*}
\begin{center}
\includegraphics[width=1.0\linewidth]{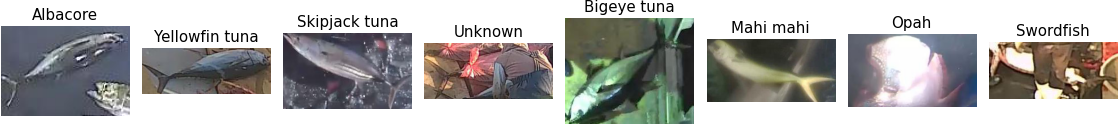}
\caption{The 8 most common L1 fish species in Fishnet. Examples selected at random and cropped by bounding box.}
  \end{center}
\label{fig:speciesex}
\end{figure*}

Large open image datasets such as ImageNet \cite{imagenet} and COCO \cite{coco} have supported the development of computer vision algorithms in other domains that can match or exceed human performance on visual recognition tasks. However, these datasets contain little or no fisheries data. Existing computer vision datasets for fish detection and species classification predominantly consist of underwater imagery sourced from ecological surveys \cite{ds10, ds5, ds8, ds0, ds6, ds7, ds12, ds11, ds1, ds4, ds9, ds3, ds2}, however underwater imagery exhibits very different characteristics from EM imagery, which is captured above water (See Figure \ref{fig:1}). Some progress has been made in developing specialized algorithms for tracking and classification of fish in EM video \cite{em5, em6, em4, em1, em0,  em2, em3}, however the data used in these studies is often difficult to access and many techniques rely upon additional specialized hardware. As of this writing, the only existing public EM data has been released through online machine learning competitions and is limited in size and diversity \cite{kaggle, dd}.

As a step toward addressing these issues, we introduce the Fishnet Open Images Database, a large dataset for detection and fine-grained visual categorization of fish species onboard commercial fishing vessels. The current release of the dataset (version 0.3) consists of 406,463 bounding boxes in 86,029 images sourced from 73 different electronic monitoring cameras, making it the largest and most diverse public dataset of fisheries electronic monitoring imagery to-date. Version 0.3 is limited to imagery from a single fishery in a relatively large geography (longline tuna in the western and central Pacific Ocean), however the challenges in this fishery are emblematic of many other large-scale industrial fishing operations around the world.

In this extended abstract we outline the data collection methodology (Section \ref{2.1}), describe the key characteristics and challenges of the dataset (Section \ref{2.2}), perform benchmarking of existing algorithms for fish detection and species classification (Section \ref{3}), and detail plans for future dataset development (Section \ref{4}).

\section{The Fishnet Open Images Database}

\subsection{Collection Methodology}
\label{2.1}
The complex and fragmented nature of the global fishing industry, coupled with evolving regulations around data confidentiality, privacy and ownership, make the collection and public release of EM data challenging. Ownership of fisheries data is typically shared between vessel operators and fisheries management agencies, who are bound by privacy contracts that restrict distribution of raw data. 

In order to assemble this dataset, we negotiated agreements with management authorities and EM service providers to obtain raw video as well as high-level catch annotations. These annotations consisted of timestamps and species classifications of catch events. We used these timestamps to extract video segments containing fishing activity, associating each segment with the species label provided. We then sampled these video segments at one frame per second, noting that each extracted image was limited to a single species classification due to the structure of the annotations. Bounding box labels were provided by the data annotation company Sama. Images with more than one fish detection were manually reviewed by domain experts to account for errors resulting from the original image-level species labels.

We took additional steps to ensure the dataset does not include any personal identifiable information, blurring human faces and excluding any camera angles that reveal unique vessel information or hull numbers.

\subsection{Dataset Characteristics}
\label{2.2}

\begin{figure}[t]
\begin{center}
\end{center}
\includegraphics[width=1.0\linewidth]{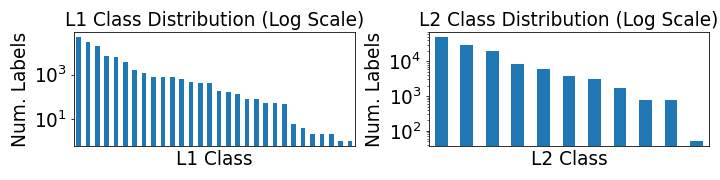}
\caption{Number of labels per fish class for the L1 and L2 label sets in Fishnet. Both distributions are long-tailed.}
\label{fig:speciesdist}
\end{figure}

\begin{figure}[t]
\begin{center}
\includegraphics[width=1.0\linewidth]{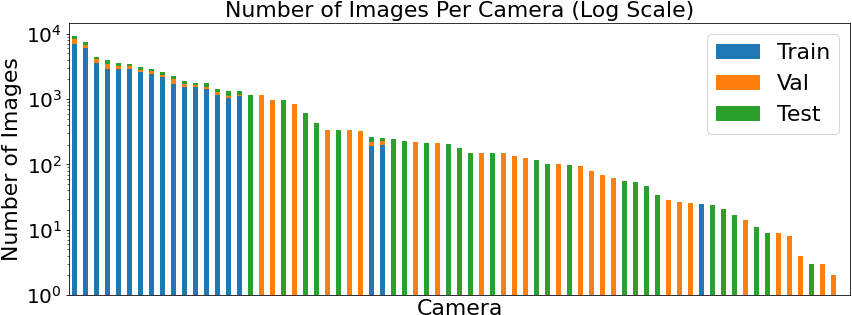}
\end{center}
   \caption{Number of images from each camera in Fishnet, and their assignment to the training, validation, and test sets. The validation and test sets contain images from cameras which \emph{do} appear in the training set as well as cameras which \emph{do not} appear in the training set, and the image distribution among cameras is long-tailed.}
\label{fig:cameras}
\end{figure}

Currently, Fishnet is limited to imagery from longline tuna vessels in the western and central Pacific, thus four visually-similar tuna species (albacore, yellowfin, skipjack, and bigeye) make up the vast majority (over 85\%) of the included fish annotations. The remainder of fish annotations are split between 25 additional species. These 29 fine-grained classes make up our ``L1'' label set. We also group related species into a set of 12 coarser classes based on the FAO ASFIS List of Species for Fishery Statistics Purposes \cite{fao}, which defines logical groupings of fish species (\eg billfish, marlin, sailfish) based on physical morphology. These classes make up our ``L2'' label set. We additionally add an ``OTH'' (Other) class at the L2 level which contains all fish labeled ``Unknown'' at the L1 level as well as any L1 classes which contain fewer than 1000 labels, excluding sharks due to their conservation importance. Humans are also annotated, as their presence and position in frame can serve as useful information for automated processing systems (\eg as an indicator of fishing activity).

The dataset is sourced from real-world fishing trips, where the distribution of encountered species is skewed. As a result, both the L1 and L2 class distributions are long-tailed, as shown in Figure \ref{fig:speciesdist}. The dataset also reflects that the distribution of catch among vessels is imbalanced, and as a result the distribution of images among cameras is also long-tailed, as shown in Figure \ref{fig:cameras}. How to deal with these data imbalances is a key challenge for the development of computer vision algorithms for EM.

Fish species can be hard to identify in EM imagery, even for expert reviewers. This is made even more challenging by the harsh operating conditions at sea, which can reduce visibility and obscure important parts of the image. Common challenges include inclement weather, water on the lens, poor lighting, fishing activity taking place at night, and chaotic crew activity which occludes fish in the frame (sometimes intentionally). See Figure \ref{fig:1} for an example. Due to these challenges, some catch events in the dataset were too difficult to classify at the species level. As a result, both the L1 and L2 label sets contain a number of ambiguous labels (\eg ``Marlin'', ``Tuna'', and ``Unknown'' in the L1 set, and ``TUNA'' and ``OTH'' in the L2 set).

\subsection{Data Split}

We construct training, validation, and test sets to mimic operating conditions in the real world. Algorithms deployed for use in EM will need to perform well on both vessels which have already been seen during training (\eg existing members of an EM program) as well as on previously-unseen vessels (\eg new members of an EM program). To this end, we draw inspiration from other datasets which source images from different unique locations \cite{iwild18, iwild19, cct, inat} and construct the Fishnet validation and test sets such that they contain approximately equal portions of imagery from cameras which \emph{do} appear in the training set (“Seen” cameras) as from cameras which \emph{do not} appear in the training set (“Unseen” cameras). The final split contains 59,497 training images, 13,648 validation images, and 12,891 test images. This distribution is depicted in Figure \ref{fig:cameras}.

\section{Experiments}
\label{3}

In order to provide an idea of the relative difficulty of Fishnet, we benchmark several common computer vision algorithms for object detection and image classification. For these experiments we include only fish labels, and we exclude ambiguous classes at the L1 and L2 levels as well as any classes with fewer than 5 labels in the training set (these classes typically appeared in just one sequence of images, and thus are not present in the test set). For all experiments, we train on 8 NVIDIA V100 GPUs and use early stopping \cite{bengio}, selecting the best model based on overall validation set performance and reporting its performance on the test set.

\subsection{Detection}

\begin{table}
\begin{center}
\begin{tabular}{|c|c|c|c|c|}
\hline
Dataset & Label Set & Classes & AP & CA-AP \\
\hline
\hline
COCO & All & 80 & 40.4 & 44.1 \\
& Single-Class & 1 & 45.4 & 45.4 \\
\hline
Fishnet & L1 & 21 & 21.3 & 46.7 \\
& L2 & 10 & 29.0 & 46.1 \\
& Tuna/Not-Tuna & 2 & 41.2 & 48.2 \\
& Fish & 1 & 48.8 & 48.8 \\

\hline
\end{tabular}
\end{center}
\caption{RetinaNet-ResNet101-FPN performance on Fishnet vs. COCO for different label sets. CA-AP means ``class-agnostic average precision,'' in which evaluation does not take class labels into account.}
\label{table:det1}
\end{table}

\begin{table}
\begin{center}
\begin{tabular}{|c|c|c|c|}
\hline
Label Set & Classes & AP-Seen & AP-Unseen \\
\hline
\hline
L1 & 21 & \textbf{25.4} & 17.1 \\
L2 & 10 & \textbf{33.8} & 22.5 \\
Tuna/Not-Tuna & 2 & 41.7 & \textbf{44.5} \\
Fish & 1 & 46.6 & \textbf{53.0} \\
\hline
\end{tabular}
\end{center}
\caption{RetinaNet-ResNet101-FPN performance on the ``Seen'' and ``Unseen'' portions of the Fishnet test set.}
\label{table:det2}
\end{table}

\begin{table*}
\begin{center}
\begin{tabular}{|c|c|c|c|c|c|c|}
\hline
Label Set & Classes & Top-1 & Top-1 Tuna & Top-1 Non-Tuna & Top-1 Seen & Top-1 Unseen \\
\hline
\hline
L1 & 21 & 73.2 & 79.9 & 41.5 & 80.3 & 62.4 \\
L2 & 10 & 75.7 & 80.9 & 48.7 & 84.0 & 63.3 \\
\hline
\end{tabular}
\end{center}
\caption{Inception-V3 top-1 species classification accuracy on Fishnet test set.}
\label{table:class}
\end{table*}

For our object detection experiments, we choose a RetinaNet \cite{retinanet} with a ResNet-101 backbone \cite{resnet} and Feature Pyramid Network \cite{fpn} pre-trained on COCO. All reported results use COCO-style AP (mAP@IOU=[.50:.05:0.95]) \cite{coco}. We train using a batch size of 16, a base learning rate of 0.0025, stochastic gradient descent with a momentum of 0.9, and focal loss with $\gamma = 2.0$. We use random horizontal flipping ($p = 0.5$) as data augmentation and train for 18 epochs total, reducing the learning rate by a factor of 10 after epochs 12 and 16. For all other hyperparameters and training settings we use the defaults from \cite{detectron2}. 

To give an idea of the baseline performance of this model architecture, we report performance of the same RetinaNet configuration on the 2017 COCO validation set. We also train a single-class model on COCO using the default settings from \cite{detectron2} by grouping all foreground objects into a single class. We report this model’s performance on the 2017 COCO validation set with the same class grouping.

For object detection on Fishnet, we report results from 4 different models, each trained on a different set of class labels: L1 fish species (21 classes), L2 fish species (10 classes), ``Tuna/Not-Tuna'' (2 classes), and ``Fish'' (1 class). Results are shown in Table \ref{table:det1}. From these initial results we get an overall idea of the difficulty of object detection in Fishnet. Performance on the fine-grained L1 and L2 label sets is notably poor, achieving only 21.3 and 29.0 AP, respectively. We notice, however, that as we group similar fish species into coarser label sets, overall detection performance improves significantly, matching observed trends in other fine-grained datasets \cite{inat}.


Motivated by these results, we also evaluate all models in a class-agnostic setting by disregarding class labels at test time in order to get an idea of object localization performance. We report these evaluation results as CA-AP (``class-agnostic AP'') in Table \ref{table:det1}. We observe a similar trend of improving performance in this metric as models are trained with coarser labels, however the magnitude of this improvement is much less significant than in the overall AP. This suggests that the poor overall AP of models trained with the L1 and L2 label sets is due to classification difficulty rather than localization difficulty.

In Table \ref{table:det2}, we compare performance on the ``Seen'' versus ``Unseen'' portions of the test set. At the L1 and L2 levels the models significantly underperform on previously-unseen cameras compared to previously-seen cameras. Considering the results from prior work on domain adaptation in static cameras \cite{cct}, these results match expectations. However, interestingly, we do not notice the same discrepancy in the models trained on our 1-class and 2-class label sets, which actually perform better on previously-unseen cameras in the test set. This suggests that the task of classification may present a greater challenge to object detectors than the task of localization when adapting to new environments. We leave further study of this trend for future work.

\subsection{Classification}

We also train and evaluate an image classifier on cropped bounding boxes of fish to illustrate the difficulty of fine-grained visual categorization of the species included in Fishnet. For these experiments we use an Inception-V3 model \cite{inception} pre-trained on ImageNet. We use an input resolution of 299 x 299 pixels, a batch size of 1024, Adam \cite{adam}, and a cyclical learning rate schedule lasting 10 epochs with a maximum learning rate of 0.001 \cite{smith}. For data augmentation we use random horizontal flipping ($p = 0.5$), random rotation ($p = 0.75$), random zoom ($p = 0.75$), and random lighting and contrast changes ($p = 0.75$).

We report the top-1 test accuracy of two models, one trained on L1 labels and one trained on L2 labels, in Table \ref{table:class}, noting that the same Inception-V3 architecture achieves a top-1 accuracy of 94.4\% on the ILSVRC 2012 validation set \cite{imagenet} and 64.2\% on the iNaturalist 2017 dataset \cite{inat}. We also report performance on 2 additional subsets of the test set: one which contains only tuna species, and one which contains all other non-tuna species. As in detection, we additionally evaluate each of these models on the ``Seen'' and ``Unseen'' portions of the test set separately. 

From these results we notice several interesting trends. First, despite the L1 label set containing more than twice as many classes, classification performance is only slightly worse. This is likely due to the fact that tuna make up over 85\% of the dataset, and that all 4 tuna species (albacore, yellowfin, bigeye, and skipjack) have their own classes at both the L1 and L2 levels. This can be confirmed by observing the performance of both models on the tuna and non-tuna subsets individually: classification performance on non-tuna species differs by more than 7\% between the two models, whereas classification performance on tuna species differs by only 1\%. For both models, this also indicates that classification in the long tail of the Fishnet species distribution (\ie non-tuna classes) is significantly more challenging than in the head of the distribution (\ie tuna classes). Finally, we see that classification on previously-seen cameras outperforms previously-unseen cameras by a large margin, matching the trend indicated in our detection experiments.

\section{Conclusions and Future Work}
\label{4}

We present the Fishnet Open Images Database, a dataset for fish detection and fine-grained visual categorization in fisheries electronic monitoring. The dataset presents challenges to existing computer vision algorithms and can serve as a benchmark for research in fine-grained categorization, long-tailed distributions, domain adaptation, and detection in low-visibility conditions. Advancements in these areas can make fisheries electronic monitoring feasible at scale and help promote sustainable fishing practices worldwide.

Future improvements will include additional annotations which will allow for the evaluation of additional tasks, such as: unique identification of image sequences to allow for sequence-level metrics; multiple-object tracking labels to allow for low-frame-rate tracking, fish re-identification, and instance-level metrics; and inclusion of rare events such as endangered, threatened, and protected species interactions for few-shot learning. We are also currently processing 100,000 additional images from the western and central Pacific to be released later this year. In the future, the goal is for Fishnet to expand to different fishery types and locations as well, and we welcome data submissions from other fisheries. 

{\small
\bibliographystyle{ieee_fullname}
\bibliography{cvpr}
}

\end{document}